\documentclass[letterpaper, 10 pt, conference]{ieeeconf} \usepackage{graphicx} 
\usepackage{caption}
\usepackage{booktabs}
\usepackage{multirow}
\usepackage{amsmath}
\usepackage{floatrow}
\usepackage{newfloat}
\usepackage{dingbat}
\usepackage{amssymb}
\usepackage{pifont}
\usepackage{subcaption}
\usepackage[ruled,vlined]{algorithm2e} 
\usepackage{tikz}
\usepackage{hyperref}
\hypersetup{
    colorlinks=true,
    linkcolor=blue,
    filecolor=magenta,
    urlcolor=cyan,
    pdftitle={Overleaf Example},
    pdfpagemode=FullScreen,
    }

\IEEEoverridecommandlockouts                              

\title{\LARGE \bf
MarineFormer: A Spatio-Temporal Attention Model for USV Navigation in Dynamic Marine Environments}

\author{Ehsan Kazemi $^{1}$, Dechen Gao $^{1, 2}$, Iman Soltani$^{1}$
\thanks{$^{1}$Laboratory for AI, Robotics and Automation, Department of Mechanical and Aerospace Engineering, University of California-Davis, CA 95616 USA.
$^{2}$Computer Science Department, University of California-Davis.
Corresponding author: Iman Soltani-{\tt\small isoltani@ucdavis.edu}.}%
}

\newcommand{\Expect}{\textstyle\mathop{\bf E}}

\begin{document}

\maketitle
\thispagestyle{empty}
\pagestyle{empty}

\begin{abstract}
Autonomous navigation in marine environments can be extremely challenging, especially in the presence of spatially varying flow disturbances and dynamic and static obstacles. In this work, we demonstrate that incorporating local flow field measurements fundamentally alters the nature of the problem, transforming otherwise unsolvable navigation scenarios into tractable ones. However, the mere availability of flow data is not sufficient; it must be effectively fused with conventional sensory inputs such as ego-state and obstacle states. To this end, we propose \textbf{MarineFormer}, a Transformer-based policy architecture that integrates two complementary attention mechanisms: spatial attention for sensor fusion, and temporal attention for capturing environmental dynamics. MarineFormer is trained end-to-end via reinforcement learning in a 2D simulated environment with realistic flow features and obstacles. Extensive evaluations against classical and state-of-the-art baselines show that our approach improves episode completion success rate by nearly 23\% while reducing path length. Ablation studies further highlight the critical role of flow measurements and the effectiveness of our proposed architecture in leveraging them. Code, videos, and supplementary materials are available at:
\url{https://soltanilara.github.io/MarineFormer/}.
\end{abstract}

\section{Introduction}

Uncrewed Surface Vessels (USVs) are becoming increasingly vital to maritime operations such as hydrographic surveying, environmental monitoring, infrastructure inspection, maritime security, and autonomous cargo transport~\cite{kumar2025designimplementationdualuncrewed}. Many of these applications take place in challenging, high-flow environments. One particularly important use case is bathymetry, which provides high-resolution underwater depth data essential for navigation safety, hydrological analysis, and infrastructure monitoring. These surveys are frequently conducted in regions with strong currents. For instance, bridge crossings where regular infrastructure monitoring is critical, are especially vulnerable to high-flow conditions, where waterway narrowing accelerates currents and increases the risk of scour, a leading cause of bridge failure. If undetected, scour can severely compromise the structural integrity of transportation infrastructure. A tragic example occurred in 1987 when the Schoharie Creek Bridge in New York collapsed due to undetected scour, resulting in multiple fatalities~\cite{schoharieBridgeCollapse}. Other examples of high-flow scenarios requiring USV-based inspection include dam water intakes and industrial wastewater discharge monitoring.

\begin{figure}
    \centering
    \includegraphics[width=1\textwidth]{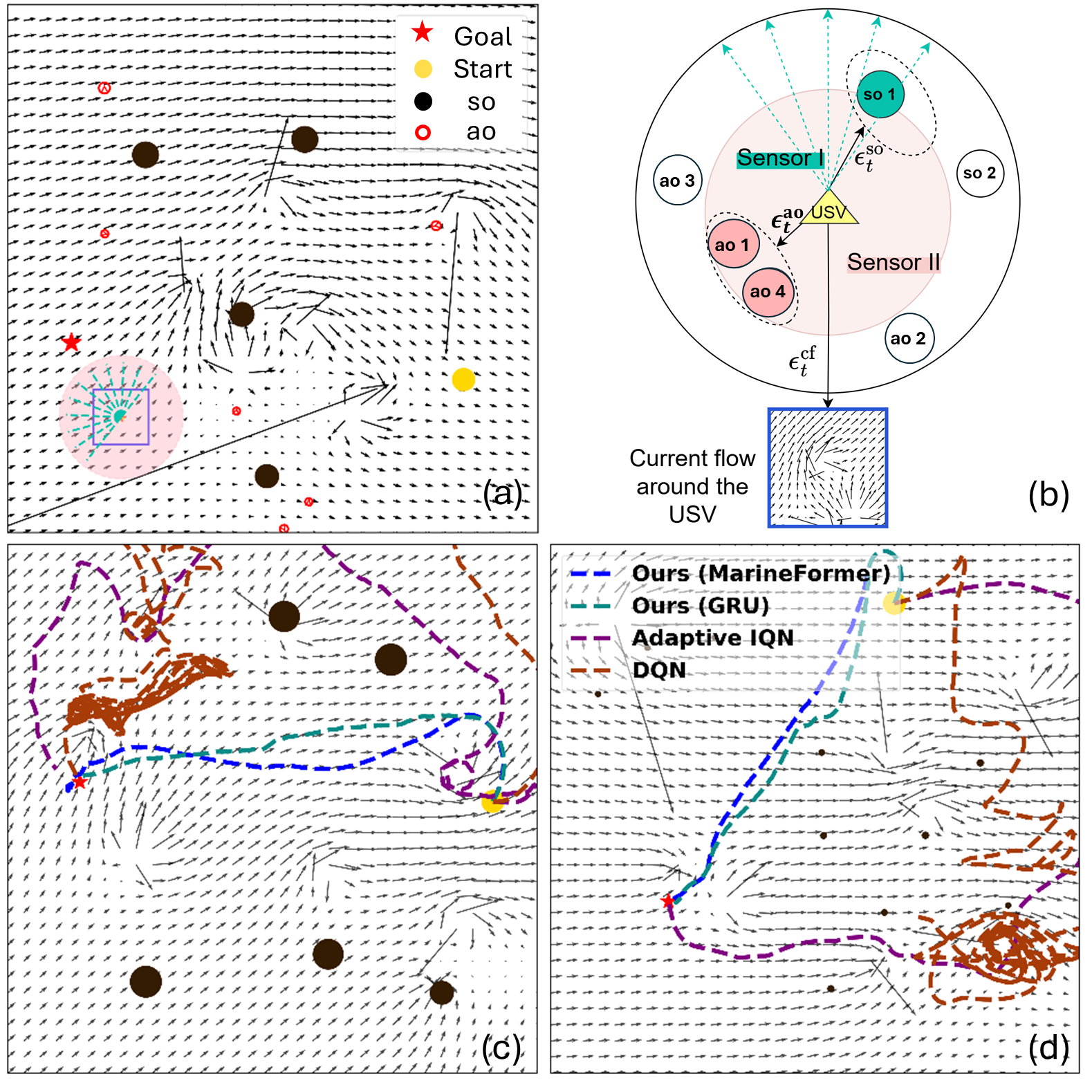}
    \caption{(a) Environment, (b) Graph model, (c, d) Example trajectories (only static obstacles are shown).}
    \label{fig:feature_input}
\end{figure}

Strong currents can significantly disrupt a USV’s trajectory and degrade the quality of bathymetry measurements. Additionally, such environments often contain dynamic obstacles, such as moving debris or other vessels, as well as static structures like bridge piers. Given the risks associated with high-flow conditions, advanced control strategies are essential for reliable USV navigation. While prior research has primarily focused on maneuvering around static and dynamic obstacles, it has largely overlooked the critical impact of flow disturbances on USV control or has treated them as unknown external forces. This gap is particularly important, as the low inertia of USVs makes them highly susceptible to flow-induced deviations. Although similar disturbances, such as wind gusts, have been studied in the context of uncrewed aerial vehicle (UAV) navigation, a key distinction is that, in USV applications, it is fundamentally possible to measure such disturbances in the nearby environment around the vehicle, enabling anticipation of upcoming high-flow regions. Techniques such as vision-based optical flow~\cite{khalid2019optical} and Doppler velocity sensing~\cite{9013073} offer viable means for estimating local flow fields, providing valuable input for navigation decisions. We argue that access to such measurements is essential for making navigation feasible in these challenging conditions. Yet, their inclusion introduces a new problem: how to effectively integrate flow information with conventional state measurements in learning-based control policies.

We tackle this challenge by introducing a reinforcement learning-based control framework that directly incorporates local flow field measurements, a conventionally missing but crucial input for safe and efficient navigation in high-flow conditions. We show that this additional sensory input transforms otherwise unsolvable navigation scenarios into tractable ones. Moreover, we demonstrate that the utility of flow data depends on how well it is fused with other sensory modalities, such as ego-state and obstacle states. To that end, we propose a novel Transformer-based architecture, \textbf{MarineFormer}, which integrates spatial attention for sensor fusion and temporal attention to model environmental dynamics.

The main contributions of this work are: 1) For the first time, we propose incorporating measurements of local flow distributions around the USV, demonstrating their significant impact on enhancing path planning capabilities. 2) We show that to fully benefit from flow measurements, effective fusion with other sensory modalities is essential. 3) We introduce MarineFormer, a novel architecture combining spatial attention and transformers that efficiently fuses multiple sensory inputs, including local flow measurements, ego-state, and obstacle information, thereby achieving significant performance improvements. 4) We show that a reward structure encouraging consistent, incremental progress toward the target, in addition to goal-reaching and obstacle avoidance, improves navigation performance. 5) We provide extensive experimental validation using both on-policy and off-policy RL algorithms. Our results demonstrate MarineFormer's robust performance, achieving nearly 23\% higher episode success rates (SR) while simultaneously reducing path length (PL).

\section{Related Works}

Path-planning algorithms for USVs can be categorized into pre-generative, reactive, and learning-based methods. Pre-generative methods, such as the Fast Marching Method (FMM) \cite{9632400}, find safe paths through known obstacles while optimizing for path length or energy consumption. Reaction-based methods, like Artificial Potential Fields (APF) \cite{fan2020improved}, generate force fields for obstacle avoidance but suffer from local minima issues. Optimal Reciprocal Collision Avoidance (ORCA) \cite{10504974} improves velocity-based collision avoidance in multi-agent settings by incorporating reciprocal obstacle avoidance but does not consider environmental disturbances. Learning-based approaches aim to overcome these limitations by making adaptive decisions \cite{liu2023intention,10804093}.

The use of transformers in RL has grown rapidly, though it introduces design complexities due to RL’s dynamic nature. In \cite{zambaldi2019deep}, self-attention mechanisms enhance relational reasoning over state representations. However, \cite{melo2022transformers} shows that standard transformers struggle with temporal sequences, sometimes performing worse than random policies. Gated Transformer-XL (GTrXL) \cite{parisotto2020stabilizing} improves stability in RL by incorporating gating layers and Identity Map Reordering, enhancing gradient flow. In this work, we employ transformers as temporal encoders, similar to \cite{banino2021coberl}, which uses temporal encoding for supervised tasks.

Path planning in dynamic flow environments has been explored in various ways. \cite{ma2018multi} formulates a multi-objective nonlinear optimization with flexible constraints. \cite{meng2022anisotropic} proposes an anisotropic GPMP2 method using an energy-based likelihood function for USV motion planning but struggles in high-density or highly dynamic environments. \cite{lan2021improved} enhances RRT by incorporating motion constraints and current-aware velocity adjustments, though it relies on predictable ocean currents. \cite{sun2017multiple} computes time-optimal trajectories via level-set equations but is restricted to predefined flow fields. \cite{xi2022comprehensive} optimizes USV performance by incorporating ocean data and physical constraints but lacks adaptability in data-scarce environments requiring real-time decision-making.

Unlike prior approaches, our method adapts to dynamic marine environments by leveraging spatio-temporal attention and local flow field measurements. While earlier works often ignore flow or treat it as unknown disturbance, we show that access to local flow data can make otherwise intractable navigation problems solvable, provided it is effectively fused with conventional sensory inputs. To this end, we use spatial attention for multimodal sensor fusion and a transformer-based encoder to model temporal dynamics, enhancing decision-making in complex conditions.



\section{Methodology}
\subsection{Environment Model}

Figure~\ref{fig:feature_input}(a) illustrates the modeled environment, where black arrows indicate the direction and magnitude of the current flow. The simulated currents are spatially varying, and incorporate singularities such as vortices, sinks, and sources. Vortices create circulating currents, sinks induce strong attraction, and sources produce repulsion.

Sources represent localized flow divergence, found at river confluences, industrial discharges, wastewater facilities, and geothermal vents \cite{8394316}. Sinks represent flow convergence, seen near hydroelectric dam intakes, spillways \cite{7401929}, drainage outflows in ports, and flood control channels. Their radial flow velocity at a distance $r$ is given by:
\begin{equation}\label{eq:SinkSource}
    V_r^{\textrm{sink/source}}=\Lambda/(2{\pi}r)
\end{equation}

Vortices, commonly found near bridge pylons, piers, and breakwaters, arise from flow separation and turbulence, and contain concentrated rotational flow \cite{A.1991}. The tangential vortex flow velocity at distance $r$ is:
\begin{equation}\label{eq:Vortex}
    V_\theta^{\textrm{vortex}}=\Gamma/(2{\pi}r)
\end{equation}

Submerged, so, and above-water, ao, obstacles are represented as black and red circles, respectively. While adaptable to any sensory modality, our framework integrates two sensors for proof of concept: sensor I (e.g., sonar) for underwater detection and sensor II (e.g., LiDAR or camera) for above-water obstacle detection. To highlight the role of spatial attention in sensory fusion, we assume sensor I detects only static submerged obstacles (so) such as riverbeds, seabeds, rocks, and snags. Sensor II identifies both static and dynamic above-water obstacles (ao), including floating debris, vessels, bridge piers, and tree branches. This setup reflects practical considerations, where at least one sonar sensor is angled downward to detect keel obstructions and shallow areas \cite{kumar2025designimplementationdualuncrewed}.

Sensor I has a 2D forward-facing field of view (FoV) (green dashed lines in Fig.~\ref{fig:feature_input}(a)), while sensor II provides 360-degree coverage (light red circle). This aligns with practical constraints, as USVs operate within a horizontal plane and obstacles positioned significantly above or below this plane are typically irrelevant to navigation \cite{kumar2025designimplementationdualuncrewed}. Additionally, a flow sensor provides a rectangular map of the local current velocity field (purple box), which in practice can be obtained using Doppler sensors \cite{9013073} or vision-based methods \cite{khalid2019optical}.

All sensor measurements are captured in the USV's local coordinate frame, except ego position and speed (e.g., from GPS), which are in the global frame. As shown in Fig.~\ref{fig:feature_input}(b), the environment is modeled as a graph with nodes representing the \textbf{USV}, \textbf{ao}, \textbf{so}, and the current flow (\textbf{cf}). Edges in this graph represent the attention of USV to observed environmental elements.

Figures~\ref{fig:feature_input} (c) and (d) show example trajectories generated by the proposed method, compared to those produced by baseline and benchmark approaches.

\subsection{USV Dynamics and Control Framework}

USV dynamics are governed by a mix of linear and nonlinear effects, shaped by hydrodynamic interactions with the environment \cite{Browning1991}. Extensive studies in classical mechanics and hydrodynamics enable accurate modeling of USV behavior under control inputs and disturbances \cite{BoatDynamics}. State-of-the-art control methods, including nonlinear \cite{ELHAKI2021108987}, optimal/adaptive \cite{9632614}, and robust \cite{Qin2022} control, provide the stability, performance, and safety guarantees essential for mission-critical USV operations.

Our RL framework assumes classical control methods handle low-level execution, eliminating the need for RL to re-learn well-understood dynamics. This simplification reduces training complexity, allowing the RL agent to focus on high-level navigation while ensuring generalizability across USV systems. This hierarchical approach has been successfully applied in other domains such as robotic manipulation and autonomous driving \cite{10433735,chuang2024activevisionneedexploring, lee2024interact}.

With low-level control managed by classical techniques, the RL policy operates within physical and control constraints, primarily defined by upper bounds on longitudinal and angular acceleration and velocity. To ensure feasible actions, we explicitly incorporate these limits in our RL formulation.

Abstracting higher-order dynamics, we model the USV using first-order equations in the presence of current \cite{lolla2014time}:
\begin{equation}
\frac{d{P_t^\textrm{ego}}}{dt} = {V_t^\textrm{ego}} = {V^{\textrm{cf}}_{P_t^\textrm{ego}}} + {V}^{\textrm{steer}}_t
\end{equation}
where $V_t^\textrm{ego}$ is the USV velocity in world coordinates, ${V}^{\textrm{steer}}_t$ is the steering velocity relative to the local current flow, and $V^{\textrm{cf}}_{P_t^\textrm{ego}}$ is the current flow velocity at the USV position $P_t^\textrm{ego}$. We assume differential thrust for the simulated USV, with actions defined as ${\bf a}_t = (\Delta v_t, \Delta \theta_t)$, where $\Delta v_t$ and $\Delta \theta_t$ represent changes in steering speed and heading, respectively. The navigation policy is given by $\pi: s_t \to (\Delta v_t, \Delta \theta_t)$.

The continuous action space is limited to $\Delta v_t, \Delta \theta_t \in [-0.1, 0.1]$ to reflect USV thruster and control limits and cap longitudinal acceleration and angular velocity. The control time step is $\Delta t = 0.25s$. We have:
\[
v^\textrm{steer}_{t+1} = v^\textrm{steer}_{t} + \Delta v_t
\]
\[
\theta^{\textrm{steer}}_{t+1} = \theta^{\textrm{steer}}_{t} + \Delta \theta_t
\]
where $v^\textrm{steer}_{t}=\|{V}^{\textrm{steer}}_t\|$, allowing the steering velocity vector to be computed as:
\[
{V}^{\textrm{steer}}_{t+1} = [v^\textrm{steer}_{t+1}\cos(\theta^{\textrm{steer}}_{t+1}), v^{\textrm{steer}}_{t+1}\sin(\theta^{\textrm{steer}}_{t+1})].
\]

\subsection{RL Formulation}

We formulate navigation as a Markov Decision Process (MDP) defined by the tuple $\mathcal{<S, A, P, R, \gamma, S}_{0}>$, where $\mathcal{S}$ and $\mathcal{A}$ denote state and action spaces, and $\mathcal{P}(s'|s, a)$ represents the state transition function. The agent selects an action $a_t$ based on state $s_t$, transitioning to $s_{t+1}$ and receiving reward $r_t = \mathcal{R}(s_t, a_t)$. The objective is to maximize the expected return $R_t = \Expect[\sum_{i=t}^T \gamma^{i-t} r_i]$, where the optimal policy $\pi^*$ satisfies the Bellman optimality equation:
\[
V^{\pi^*}(s) = \Expect[r_t + \gamma\, \max_{s_{t+1}}V^{\pi^*}(s_{t+1}) | s_t=s]
\]
where $s_{t+1} \sim \mathcal{P}(\cdot~| s, a)$ and $a \sim \pi^*(s)$. A goal is reached when the Euclidean distance to the target is within $d_\textrm{tgt}$.

The policy receives state observations $s_t = [s_t^{\textrm{ego}}, s_t^{\textrm{so}}, s_t^{\textrm{ao}}, s_t^{\textrm{cf}}]$. The ego-state $s_t^{\textrm{ego}}$ consists of the USV’s position and velocity in the global coordinate:
\[
s_t^{\textrm{ego}} = [P_{t}^\textrm{ego}, V_{t}^\textrm{ego}]
\]
where $P_{t}^\textrm{ego} = [p^x_{t}, p^y_{t}]$ and $V_{t}^\textrm{ego} = [v^x_{t}, v^y_{t}]$.

The state of submerged obstacles are represented as:
\[
s_t^{\textrm{so}} = [O^{\textrm{so}_1}_{t}, \ldots, O^{\textrm{so}_{N}}_{t}]
\]
where $O^{\textrm{so}_i}_{t}$ is a measurement from the $i^\textrm{th}$ beam of sensor I, with range $M_{R_1}$. For above-water obstacles we have:
\[
s_t^{\textrm{ao}} = \left[O^{\textrm{ao}_1}_{t}, \ldots, O^{\textrm{ao}_M}_{t}\right]
\]
where $O^{\textrm{ao}_j}_{t}$ concatenates the position, velocity, and their predicted trajectory over a $K$-step horizon:
\begin{equation}\label{eq:dynamic_obs}
O^{\textrm{ao}_j}_{t} = [P^{\textrm{ao}_j}_{t}, V^{\textrm{ao}_j}_{t}, \hat{P}^{\textrm{ao}_j}_{t+1:t+K}]
\end{equation}
with $P^{\textrm{ao}_j}_{t}$ and $V^{\textrm{ao}_j}_{t}$ being the position and velocity, and $\hat{P}^{\textrm{ao}_j}_{t+1:t+K}$ the predicted trajectory. A constant velocity assumption is used for prediction, though ground-truth trajectories can be incorporated if available.

The sizes of $s_t^{\textrm{ao}}$ and $s_t^{\textrm{so}}$ depend on the beam count of sensor I and the detection capacity of sensor II, with masking applied to exclude undetected or out-of-range obstacles.

The policy also receives flow velocity field measurements $s_t^{\textrm{cf}}$ on an $m \times m$ grid surrounding the USV:
\[
s_t^{\textrm{cf}} = \{V_{t,i,j}^{\textrm{cf}}=(v^{\textrm{cf},x}_{t,i,j}, v^{\textrm{cf},y}_{t,i,j})\}_{i,j=0}^m
\]
where $i, j$ are grid indices.

\subsection{Network Architecture}

\begin{figure}
    \centering
    \includegraphics[width=0.95\textwidth]{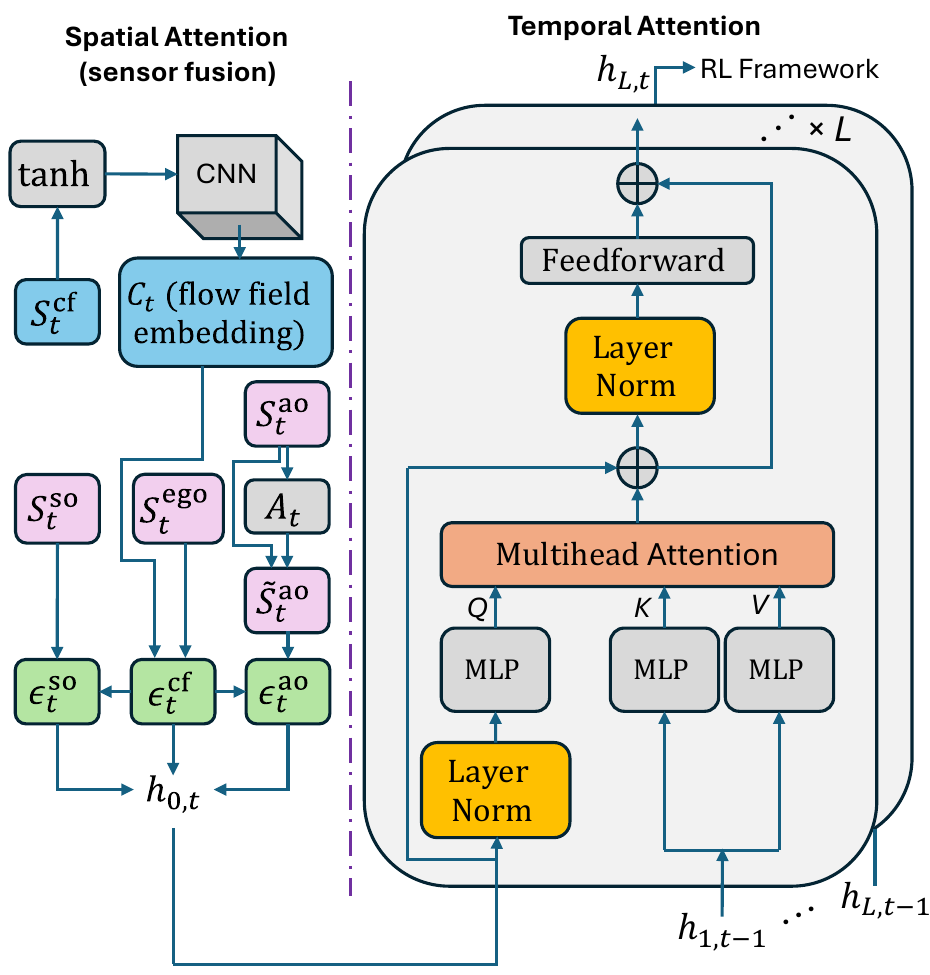}
    \caption{The architecture of the MarineFormer model.}
    \label{fig:architecture}
\end{figure}

MarineFormer integrates spatial and temporal attention for navigation. The spatial component models the USV-environment relationship as a graph (Fig. \ref{fig:feature_input}b), processing sensory inputs such as position, speed, and flow disturbances. The processed inputs are fed to a Transformer, which captures the temporal evolution of the environment (Fig. \ref{fig:architecture}).

\subsubsection{Spatial Attention to the Environment}

At each time step $t$, the spatio-temporal graph $G_t = (N_t^A, \epsilon_t^{A'})$ consists of nodes $N_t^A$, where $A$ represents USV, cf, ao, and so, and edges $\epsilon_t^{A'}$ represent the USV’s attention to environmental elements. The nodes are defined as $s_t^{\textrm{ego}}$, $s^{\textrm{cf}}_t$, $s^{\textrm{so}}_{t}$, and $\widetilde{s}^{\textrm{ao}}_{t}$, where $\widetilde{s}^{\textrm{ao}}_{t}$ is an augmented version of ${s}^{\textrm{ao}}_{t}$.

Spatial attention is computed as:
\begin{equation}\label{eq:attention}
\epsilon_t^{A'} = softmax\left(\frac{Q_t^{A'}\,{K_t^{A'}}^{T}}{\sqrt{d}}\right)\,V_t^{\textrm{attn},A'}
\end{equation}
where $d$ is the query and key dimension. For all nodes, keys and values are separate embeddings of the node. Queries, $Q_t^{\textrm{ao}}$ and $Q_t^{\textrm{so}}$, are derived from $\epsilon_t^{\textrm{cf}}$, incorporating flow disturbances in obstacle attention. $Q_t^\textrm{cf}$ is derived from $s_t^{\textrm{ego}}$.

The dynamic nature of above-water obstacles makes them a key challenge. To capture obstacle interactions, we introduce an alignment matrix:
\begin{equation}\label{eq:sim_score}
{\mathbf{A}}_t = softmax\left({\mathbf{B}}_t\mathbf{D}\mathbf{D}^T{{\mathbf{B}}_t}^T\right)
\end{equation}
where $\mathbf{D}$ is a trainable diagonal weight matrix, and,
\begin{equation}
\mathbf{B}_t = \begin{bmatrix}
{O}^{\textrm{ao}_1}_{t} \\
\vdots \\
{O}^{\textrm{ao}_M}_{t} \\
\end{bmatrix}
\end{equation}
Each ao state is augmented with corresponding row from $\mathbf{A}_t$.
\begin{equation}\label{eq:dynamic_obs2}
\widetilde{O}^{\textrm{ao}_j}_{t} = [{O}^{\textrm{ao}_j}_{t}, \mathbf{A}_{t}^{j}]
\end{equation}
where $\mathbf{A}_{t}^{j}$ is the $j^{\textrm{th}}$ row of $\mathbf{A}_{t}$. Finally, the ao node is:
\[
N_t^{\textrm{ao}}=\Tilde{s}_t^{\textrm{ao}} = \left[\Tilde{O}^{\textrm{ao}_1}_{t}, \ldots, \Tilde{O}^{\textrm{ao}_M}_{t}\right]
\]
All embeddings $Q, K$, and $V$ (except $K^\textrm{cf}$ and $V^\textrm{cf}$) are obtained via multilayer perceptrons (MLPs). The flow measurement node, $N_t^{\textrm{cf}} = s^{\textrm{cf}}_t$, is mapped to $[-1,1]$ via $\tanh$ and processed through a convolutional network, producing a latent representation $C_t$. $K^\textrm{cf}$ and $V^\textrm{cf}$, are derived from $C_t$, allowing the model to capture velocity gradients, and disturbance patterns.

\subsubsection{Capturing Temporal Dependencies through Transformer}
The graph $G_t$ evolves due to agent motion and flow disturbances. A Transformer (Fig. \ref{fig:architecture}) captures these dynamics, processing the graph attention edges:
\[
h_{0, t} = [\epsilon_t^{\textrm{cf}}, \epsilon_t^{\textrm{ao}}, \epsilon_t^{\textrm{so}}]
\]
The Transformer output, $h_{L, t}$, passes through a fully connected layer to produce the action, $a_t$, the value, $V$, or $Q$, depending on the adopted RL framework. Algorithm \ref{alg:SpatialTransformer} summarizes the MarineFormer architecture.

\subsection{Reward Function}
The reward function is defined as:
\begin{equation}\label{eq:total_reward}
\begin{split}
 & r_t  =   {\bf 1}_{\textrm{collision}}(s_t) r_{\textrm{c}} + {\bf 1}_{\textrm{goal}}(s_t) r_{\textrm{g}} \\
 & + {\bf 1}_{\textrm{collision}^c \cap \textrm{goal}^c}(s_t) \big(r_{t}^{\textrm{pot}}  + r_{t}^\textrm{ao} + r_{t}^\textrm{so} + r_{t}^{\textrm{cf}} \big)
 \end{split}
\end{equation}
where ${\bf 1}_{\textrm{collision}}$ and ${\bf 1}_{\textrm{goal}}$ are indicator functions for collision and goal arrival, respectively. A collision occurs when the USV is within $d_{\textrm{col}}$ of an obstacle, while reaching $d_{\textrm{tgt}}$ signifies arrival at the target. The penalties for collision and the rewards for reaching the goal are $r_{\textrm{c}}$ and $r_{\textrm{g}}$. The term ${\bf 1}_{\textrm{collision}^c \cap \textrm{goal}^c}$ applies when neither event occurs.

The potential reward guides the USV toward the goal:
\[
r^{\textrm{pot}}_t=  \alpha\,(d^{\textrm{goal}}_{t-1} - d^{\textrm{goal}}_{t})
\]
where $d^{\textrm{goal}}_{t}$ is the $L_2$ distance to the goal at time $t$, and $\alpha$ is a scalar hyperparameter.

To mitigate collision risks due to limited maneuverability, penalties $r^{\textrm{ao}}_{t}$ and $r^{\textrm{so}}_{t}$ discourage proximity to obstacles:
\begin{equation}\label{eq:prediction_reward}
\begin{split}
    r^{\textrm{ao}}_{t} & = \min_{i = 1, \ldots, M} \min_{k=1,\ldots, K_\textrm{ao}} (\frac{{\bf 1}_{\textrm{pred-coll}}^{i}(\hat{s}_{t+k})r_{c}}{2^{k+2}}) \\
    r^{\textrm{so}}_{t} & = \min_{i = 1, \ldots, N} \min_{k=1,\ldots, K_\textrm{so}} (\frac{{\bf 1}_{\textrm{pred-coll}}^{i}(\hat{s}_{t+k})r_{c}}{2^{k+2}}) \\
\end{split}
\end{equation}
where $M$, $N$, $K_\textrm{ao}$ and $K_\textrm{so}$ are the numbers of above- and under-water obstacles, and their corresponding collision prediction horizons, respectively. ${\bf 1}_{\textrm{pred-coll}}^{i}$ is active when the USV comes within $d_{\textrm{enc}}$ of obstacle $i$ at predicted future state $\hat{s}_{t+k}$. The encroachment radius $d_{\textrm{enc}}$ accounts for sensor noise and prediction errors, with larger values leading to more conservative behavior. Penalties decrease proportionally with prediction step $k$ to reflect increasing uncertainty.

$r^{\textrm{ao}}_{t}$ accounts for predicted collisions using both USV and dynamic obstacle's predicted trajectories, while $r^{\textrm{so}}_{t}$ considers only USV movement. The penalty only considers the most imminent projected encroachment.

To prevent stagnation in high-flow, $r^{\textrm{cf}}_{t}$ penalizes steps where the USV fails to move at least $d_{\textrm{cf}}$ toward the goal:
\vspace{-2pt}
\[
r^{\textrm{cf}}_{t} = {\bf 1}_{\textrm{stagnation}}(s_t) r_{\textrm{cf}}
\]
This encourages policies that exploit currents for efficient navigation while avoiding flow-induced traps.
\vspace{-2pt}
\begin{equation}\label{eq:r_disp}
{\bf 1}_{\textrm{stagnation}}(s_t) =
     \begin{cases}
       1 &\quad\text{if } d^{\textrm{goal}}_{t-1} - d^{\textrm{goal}}_{t} < d_{\textrm{cf}}, \\
       0 &\quad\text{otherwise.} \\
     \end{cases}
\end{equation}
\begin{algorithm}
\footnotesize
\caption{\textbf{Spatial Attention + Transformer}} \label{alg:SpatialTransformer}
\hrulefill
\begin{flushleft}
\textbf{Input:} Ego-state ($s^{ego}_t$), submerged ($s^{so}_t$), and above-water ($s^{ao}_t$) obstacles, current flow ($s^{cf}_t$). All $\Phi$ denote MLP architectures, except $\Phi_{\text{cf}}^K$ and $\Phi_{\text{cf}}^V$, which represent convolutional networks.\\
\vspace{-10pt}
\end{flushleft}
\hrulefill
\vspace{-7pt}
\begin{flushleft}
\textbf{Step 1: Graph Nodes}
\end{flushleft}
\vspace{-10pt}
\begin{align*}
    N_t^{ego} &= s^{ego}_t \:  \text{(Ego-state node)} \\
    N_t^{cf} &= s^{cf}_t \: \text{(Current flow node)} \\
    N_t^{so} &= s^{so}_t \: \text{(Static obstacle nodes)} \\
    A_t &= \text{ComputeAlignment}(s_t^{ao}) \: \text{(Alignment matrix)} \\
    N_t^{ao} &= \left[ s^{ao}_t, A_t \right] \: \text{(Augmented ao nodes)}
\end{align*}
\vspace{-17pt}
\begin{flushleft}
\textbf{Step 2: Graph Edges $G_t$ (Spatial Attention)}
\end{flushleft}
\vspace{-10pt}
\begin{align*}
    Q^{ao}_t &= \Phi_{Q}^{\textrm{ao}}(\epsilon^{cf}_t), \; Q^{so}_t = \Phi_{Q}^{\textrm{so}}(\epsilon^{cf}_t), \; Q^{cf}_t = \Phi_{Q}^{\textrm{cf}}(N_t^{ego}) \\
    K^{ao}_t &= \Phi_{K}^{\textrm{ao}}(N_t^{ao}), \; K^{so}_t = \Phi_{K}^{\textrm{so}}(N_t^{so}), \; K^{cf}_t = \Phi_{K}^{\textrm{cf}}(N_t^{cf}) \\
    V^{ao}_t &= \Phi_{V}^{\textrm{ao}}(N_t^{ao}), \;  V^{so}_t = \Phi_{V}^{\textrm{so}}(N_t^{so}), \; V^{cf}_t = \Phi_{V}^{\textrm{cf}}(N_t^{cf}) \\
\end{align*}
\vspace{-30pt}
\begin{align*}
    \epsilon^{ao}_t &= \text{Attention}(Q^{ao}_t, K^{ao}_t, V^{ao}_t), \\
    \epsilon^{so}_t &= \text{Attention}(Q^{so}_t, K^{so}_t, V^{so}_t), \\
    \epsilon^{cf}_t &= \text{Attention}(Q^{cf}_t, K^{cf}_t, V^{cf}_t)
\end{align*}
\vspace{-20pt}
\begin{align*}
    h^{0,t} &= [\epsilon^{cf}_t, \epsilon^{ao}_t, \epsilon^{so}_t] \quad
\end{align*}
\vspace{-18pt}
\begin{flushleft}
\textbf{Step 3: Temporal Attention with Transformer}
\end{flushleft}
\vspace{-10pt}
\begin{align*}
    h^{\textrm{trans}}_t &= \text{Transformer}(h_{0,t-\tau:t})
\end{align*}
\vspace{-18pt}
\begin{flushleft}
\textbf{Step 4: RL related computations using $h^{\textrm{trans}}_t$}.
\end{flushleft}
\vspace{-7pt}
\hrulefill
\end{algorithm}

\vspace{-8pt}
\section{Experiments}
\subsection{Simulation Setup}
We conduct tests under two different environments with varying obstacle counts and sizes. The first includes 10 above-water and 5 submerged obstacles (denoted hereafter as 10/5), while the second increases these limits to 25 and 10 (25/10), respectively. Static underwater obstacles in the first set have radii randomly chosen from $[1.0, 1.5]$m, whereas those in the second set are smaller, ranging from $[0.2, 0.3]$m.

Above-water obstacles are controlled by ORCA, reacting to dynamic obstacles but not the USV, preventing the agent from exploiting obstacle yielding. These obstacles regularly change velocity and direction, exposing the limitations of the RL agent’s constant-velocity assumption for encroachment prediction. Additionally, dynamic obstacles are unaffected by current flow and randomly switch assigned goals mid-episode. Unlike cooperative obstacle avoidance frameworks, our approach assumes fully independent obstacle movement, creating a more challenging scenario where the USV must react to unpredictable behaviors without coordination.

Above-water obstacles have a radius of $0.3$m, a max velocity of $2$m/s, and a $2\pi\,\text{rad}$ FoV in ORCA. The environment, a $40 \textrm{m} \times 40$m area, is denser than those in prior works \cite{10804093, lin2024decentralized}, making obstacle avoidance more challenging. While the USV is not confined within this region, leaving it increases travel time, leading to implicit penalization.

Each episode randomly initializes the USV’s start and goal positions, obstacle locations, uniform flow angle of attack, and flow singularities (vortices, sinks, and sources). The uniform flow speed is set to $V^{\textrm{uniform}} = 1$ m/s with an attack angle randomly selected from $[0, \pi/4]\,\text{rad}$. The velocity field is simulated using potential flow techniques \cite{acheson1990elementary}, incorporating four vortices and four sinks/sources per environment. The singularity strengths $\Lambda$ and $\Gamma$ (Eqs. \ref{eq:SinkSource} and \ref{eq:Vortex}) are randomly selected from $[5\pi, 10\pi]$. The total velocity field at position $p=[p^x,p^y]$ is:
\[
V^{\textrm{cf}}_p=V^{\textrm{source}}_{p}+V^{\textrm{sink}}_{p}+V^{\textrm{vortex}}_{p}+V^{\textrm{uniform}}.
\]

Sensor I uses $N = 11$ beams for submerged static obstacles, while Sensor II can track up to $M = 30$ static/dynamic above-water obstacles. We use a similar $5$m range for both sensors, though each can have a different range without affecting training. The current flow is measured on an $8 \times 8$ grid over a $5\textrm{m} \times 5$m area around the USV. Sensor configurations can be adjusted to fit specific USV setups.

\subsection{RL Training}

Since our approach is agnostic to the underlying RL algorithm, we evaluate MarineFormer with both on-policy and off-policy methods. Specifically, we use Proximal Policy Optimization (PPO) \cite{schulman2017proximal} as the on-policy baseline, and Soft Actor-Critic (SAC) \cite{haarnoja2018soft} and Twin Delayed DDPG (TD3) \cite{fujimoto2018addressing} as off-policy counterparts, three widely adopted RL algorithms with strong performance across diverse tasks \cite{arulkumaran2017deep}.

MarineFormer’s architecture remains consistent across experiments, with minor variations in components and hyperparameters. In our experiments, we consistently use one attention head and one transformer layer, though these can vary based on environment complexity and computational resources. PPO relies on a clipping parameter and entropy regularization, while SAC and TD3 use replay buffers, target networks, and algorithm-specific mechanisms like temperature tuning (SAC) or delayed updates (TD3).

To support off-policy learning, we store Transformer hidden states in the replay buffer \cite{kapturowski2018recurrent}, enabling reuse in actor, critic, and target predictions. Unlike PPO, which uses a shared Transformer backbone, off-policy methods use separate backbones for the actor/critic and target networks \cite{yang2021recurrent}, avoiding instability from mixing Polyak-averaged and backprop-trained parameters. In practice, this separation yields more stable training and improved performance.

\subsection{Evaluation}
For the reward function \eqref{eq:total_reward}, we set $r_\textrm{c} = -20$, $r_{\textrm{cf}}=-0.2$, and $r_\textrm{g}=10$. Prediction horizons $K_{\textrm{ao}}$ and $K_\textrm{so}$ in \eqref{eq:prediction_reward} are set to $5$ and $3$, respectively. Other key parameters include $d_{\textrm{cf}} = 0.07$, and $d_{\textrm{tgt}}=0.6$.

Minimum start-to-goal distance is $12$m, minimum obstacle separation is $5$m, and each episode is limited to $110$s ($440$ steps). Hyperparameter selection varies by RL scheme and is detailed on the article's GitHub page.

We evaluate MarineFormer on the two test environments and compare it to both classical and learning-based baselines. Classical methods include ORCA~\cite{van2011reciprocal}, which enforces reciprocal collision avoidance, and APF ~\cite{fan2020improved}, which uses attractive and repulsive forces to guide the USV. Learning-based baselines include Marine SAC (M-SAC) \cite{10804093}, Marine DQN (M-DQN) ~\cite{mnih2015human}, Marine Adaptive IQN (M-A-IQN) \cite{lin2024decentralized}, and its Actor-Critic counterpart M-AC-IQN \cite{10804093}. M-SAC, M-DQN and M-A-IQN/M-AC-IQN are policies based on Soft Actor-Critic (SAC), Deep Q-Networks (DQN) and Implicit Quantile Networks (IQN) ~\cite{dabney2018implicit}, respectively, implemented for navigation in dynamic marine environments. We also evaluate modified versions of these models that receive local velocity (flow) measurements along with their original sensory inputs.

\begin{figure}
    \centering
    \vspace{-10pt}
    \includegraphics[width=1.0\textwidth]{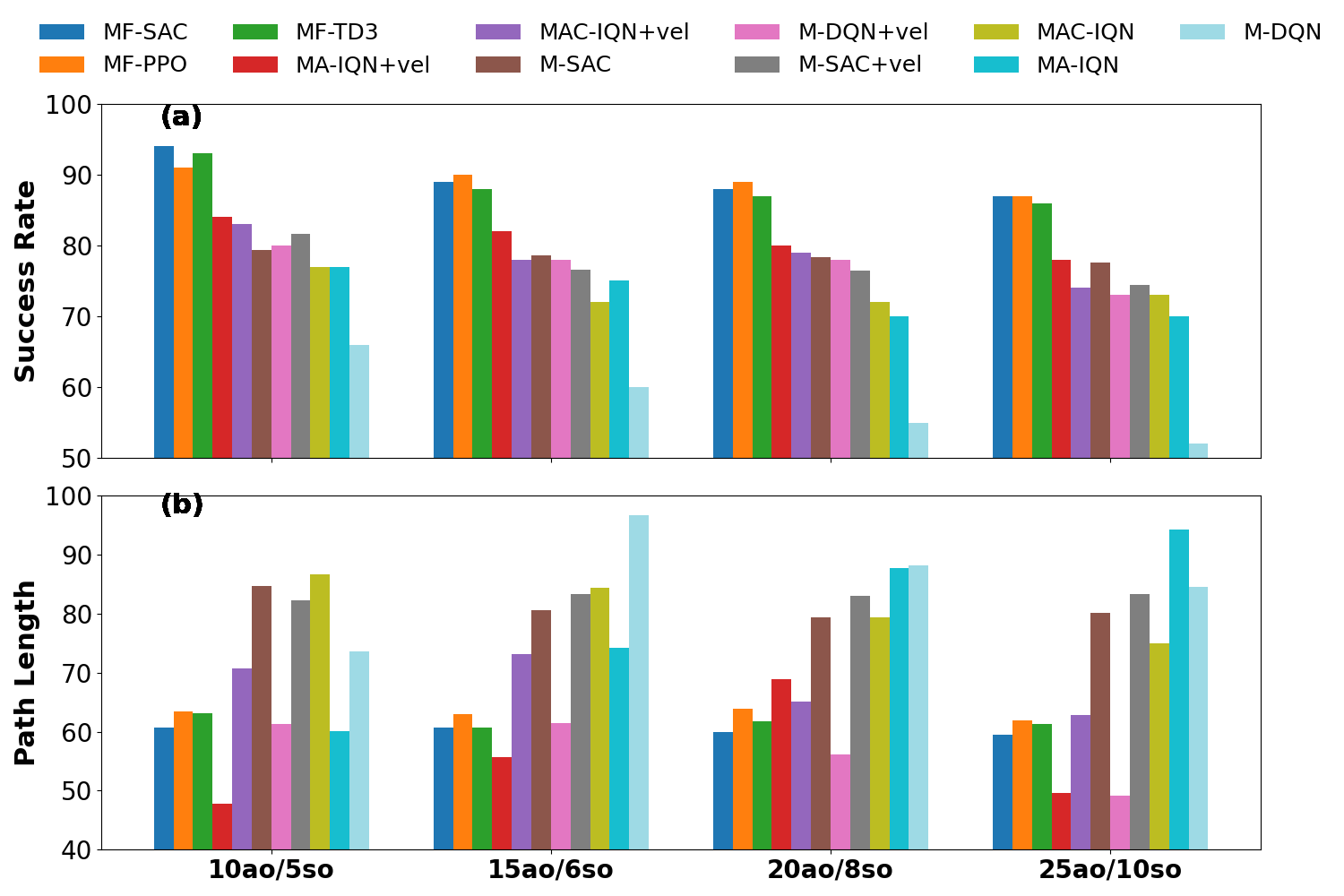}
    \caption{(a) Success rate and (b) path length, for different methods, against varying number of obstacles (ao/so).}
    \label{fig:SuccessRateVSObst}
\end{figure}

The results of table~\ref{tab:overall_model_perf} reveal several key insights:

\textbf{1) Velocity measurements improve all methods.} Across all models, adding flow velocity input boosts performance. This supports our central claim that access to flow information is crucial for effective navigation in high-disturbance environments. Notably, both M-A-IQN and M-DQN greatly benefit in terms of PL reduction.

\textbf{2) MarineFormer achieves balanced gains in both SR and PL.} While M-A-IQN and M-DQN see substantial reductions in path length, outperforming MarineFormer on that specific metric, their SRs remain significantly lower. This suggests that the immediate benefit of flow information lies in improving PL. However, simultaneously achieving higher SRs, which involve complex interactions between flow disturbances and obstacle avoidance, requires more than simply observing flow, it requires fusing it meaningfully with other state inputs and modeling longer-term dependencies. Despite the fact that our reward structure heavily penalizes collisions compared to how it rewards shorter paths, M-A-IQN and M-DQN seem unable to cross a performance threshold in SR. In contrast, MarineFormer is able to leverage flow data for both efficiency and safety, resulting in a more balanced and robust improvement across both metrics.

\textbf{3) Architecture generalizes across RL frameworks.} MarineFormer consistently delivers strong performance across PPO, SAC, and TD3, demonstrating its independence from the underlying learning algorithm. SAC achieves the highest SR (93\%) on the first test set, while PL performance across MarineFormer variants remains competitive with velocity-augmented benchmarks. Importantly, MarineFormer offers consistent and high performance across both dense and moderately dense obstacle environments.

\textbf{4) Impact of temporal modeling.} Replacing the transformer with a GRU reduces SR and increases PL, confirming that transformer-based temporal attention is better suited to capturing the long-term dependencies required for robust navigation in dynamic and disturbed marine environments.


\begin{table}
\centering
\caption{Success rate and path length results.}

\label{tab:overall_model_perf}
\resizebox{1\linewidth}{!}{%
  \begin{tabular}{|c|c|c||c|c|}
    \toprule
    \multirow{2}{*}{{\bf Method}} & \multicolumn{2}{c||}{{\bf First test set}} & \multicolumn{2}{c|}{{\bf Second test set}} \\ \cline{2-5}
     & SR & PL & SR & PL \\\cline{1-5}
    APF \cite{fan2020improved} &62\% &201.84 & 60\% & 195.23 \\\cline{1-5}
    ORCA \cite{van2011reciprocal} &7\% & 112.46 & 5\% &105.57 \\\cline{1-5}
    M-DQN \cite{mnih2015human} & 68\% & 106.98 &52\% &84.56\\\cline{1-5}
    M-A-IQN \cite{lin2024decentralized} &70\% & 60.73 &70\% &94.28 \\\cline{1-5}
    M-AC-IQN \cite{10804093}  & 75\% & 87.59 & 73\% & 74.91 \\\cline{1-5}
    M-SAC \cite{10804093}  & 81\% & 86.69 & 77\% & 86.36 \\\cline{1-5}
    M-AC-IQN \cite{10804093} + velocity  & 81\% & 77.02 & 74\% & 62.85 \\\cline{1-5}
    M-SAC \cite{10804093} + velocity  & 80\% & 82.28 & 76\% & 81.21 \\\cline{1-5}
    M-A-IQN \cite{lin2024decentralized} + velocity &82\% & 45.39 &78\% &49.66 \\\cline{1-5}
    M-DQN \cite{mnih2015human} + velocity &83\% & 48.85 &73\% &49.22 \\\cline{1-5}
    Ours (PPO+GRU) &86\% & 61.90 & 85\% & 63.86 \\\cline{1-5} 

    Ours (MarineFormer+PPO (MF-PPO)) & 91\% & 61.47 & {\bf 87\%} & {61.93} \\\cline{1-5}  

    Ours (MarineFormer+SAC (MF-SAC)) & {\bf 93\%} & 58.51 & {\bf 87\%} & 59.53 \\\cline{1-5}  
    Ours (MarineFormer+TD3 (MF-TD3)) & 92\% & 60.28 & 86\% & 61.37 \\
    \bottomrule
  \end{tabular}
  }%
\end{table}

To assess generalization, we evaluated models trained on the most challenging environment (Test Environment 2: 25/10) in progressively easier scenarios. As shown in Fig.~\ref{fig:SuccessRateVSObst}, these policies often matched or even outperformed models trained directly on the easier task (Test Environment 1: 10/5). This suggests that training under high-obstacle, high-disturbance conditions yields robust policies that transfer well to less demanding environments, supporting the practical strategy of training under an upper bound of task difficulty. We also observe that while the performance gap in PL between methods narrows in simpler scenarios, the gap in SR widens. For instance, under the 25/10 obstacle configuration, MarineFormer-SAC outperforms M-A-IQN by nearly 21\%, compared to a 17\% gap in the 10/5 environment. This is intuitive: in low-obstacle environments, the optimal path is relatively well-defined, requiring only occasional short-term deviations for obstacle avoidance. As a result, most policies complete the episode with a near-optimal PL. In contrast, denser environments demand more frequent and complex navigational adjustments. Less capable policies tend to accumulate greater deviations from the optimal trajectory, amplifying differences in path length performance. Additionally, in such settings, a single failure, such as colliding with the last of several obstacles after successfully avoiding the others, results in episode failure. This outcome is similar for a weaker policy failing at the first obstacle, thereby narrowing observable differences in SR across methods. These observations highlight the importance of evaluating both success rate and trajectory quality across a range of obstacle densities. Low-obstacle environments better reveal a model’s fundamental collision avoidance capabilities, allowing finer-grained comparisons. In contrast, high-obstacle environments provide a more rigorous test of a policy’s ability to maintain efficient trajectories while navigating persistent threats.

\begin{figure*}[thpb]
    \centering
    \includegraphics[width=1\textwidth]{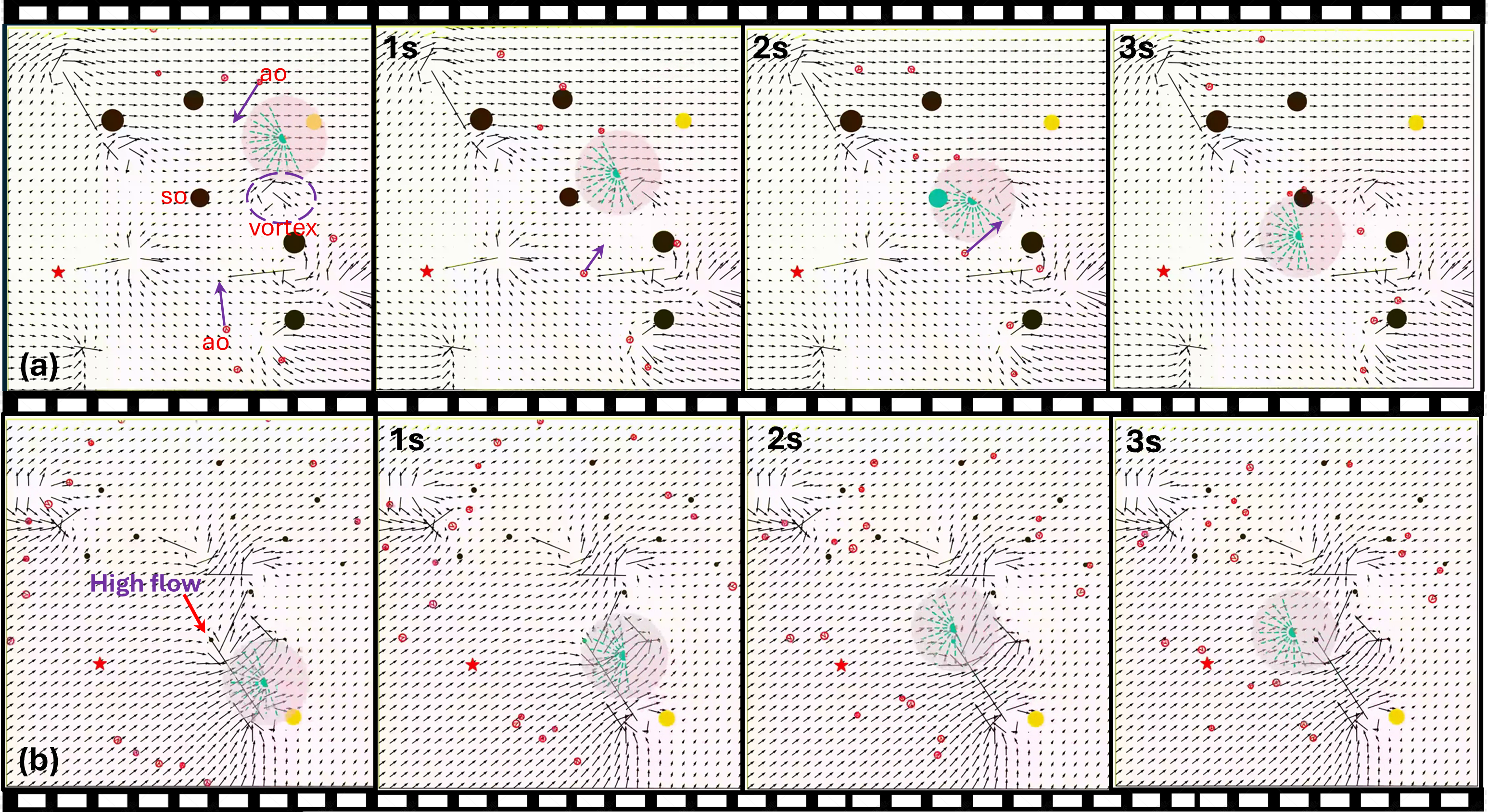}
    \caption{Consecutive frame of USV navigating complex scenarios. a) going through a vortex, a static obstacle (so) and a dynamic obstacle (ao) closing in, and b) avoiding a high-flow region.}
    \label{fig:MovieStrip}
\end{figure*}

Finally, Fig.~\ref{fig:MovieStrip} presents two challenging navigation scenarios handled by MarineFormer. In scenario (a) the USV successfully navigates through a vortex, a static and a dynamic obstacle closing in and in scenario (b) it avoids a high-flow region. These scenarios exemplify situations where baseline and benchmark methods frequently fail. Overall, the results highlight MarineFormer's ability to make \textbf{proactive, long-sighted decisions}, especially in complex marine environments with dense obstacles and strong flow disturbances.

\begin{table}
\centering
{%
  \caption{Ablation study on the components of the proposed method. Included (\checkmark), excluded (\ding{55}).}
  \label{tab:ablation_on_reward_components}%
}
\resizebox{0.8\linewidth}{!}{%
    \begin{tabular}{|c|c|c|c|c|c|c|}
    \toprule
    \multirow{2}{*}{$r^{\textrm{ao}}_{t}$} & \multirow{2}{*}{$r^\textrm{so}_{t}$} & \multirow{2}{*}{$r^{\textrm{cf}}_{t}$} & \multirow{2}{*}{Alignment} & Current flow  & \multirow{2}{*}{SR} & \multirow{2}{*}{PL}\\
    &  &  &  & measurement &  &  \\
    \midrule
    \checkmark & \checkmark & \checkmark & \checkmark & \checkmark & 91\% &61.47\\ \hline 
    \ding{55} & \checkmark & \checkmark & \checkmark & \checkmark & 71\% & 65.33\\ \hline 
    \checkmark & \ding{55} & \checkmark & \checkmark & \checkmark & 77\% & 63.95\\ \hline 
    \checkmark & \checkmark & \ding{55} & \checkmark & \checkmark & 65\% & 67.50\\ \hline 
    \checkmark & \checkmark & \checkmark & \ding{55} & \checkmark & 83\% & 62.82\\ \hline 
    \checkmark & \checkmark & \checkmark & \checkmark & \ding{55} & 71\% & 70.40\\ \hline 
    \bottomrule
    \end{tabular}
  }
\end{table}
\subsection{Ablation Study}

We analyze the impact of our method's components through ablations when PPO is used as the RL framework (Table ~\ref{tab:ablation_on_reward_components}). Results show that all reward elements $r^\textrm{ao}_{t}$, $r^\textrm{so}_{t}$, and $r^{\textrm{cf}}_{t}$ significantly influence performance, with $r^{\textrm{cf}}_{t}$ having the most prominent effect. This underscores the importance of guiding the USV not just to avoid obstacles but also to ensure steady progress toward the target. Incorporating alignment improves SR by at least $7\%$.

Table ~\ref{tab:ablation_on_reward_components} also shows that removing flow measurements significantly reduces SR and increases PL, brining the performance almost to the same level as that of the benchmarks. However, as noted earlier and observed in Table \ref{tab:overall_model_perf} upon the availability of flow measurements, MarineFormer, can make a more effective use of this data.

\section{Limitations and Future Work}

\textbf{Simulation-Only Evaluation:} Our current evaluation is limited to simulation. Future efforts will focus on field experiments. For this purpose, we have recently developed a bathymetry USV platform capable of executing high-level speed and heading commands \cite{kumar2025designimplementationdualuncrewed}.

\textbf{Dependence on 2D Flow Measurements:} The method requires local 2D flow measurements, but no commercial sensor provides this data. We plan to develop a vision-based flow estimation approach inspired by existing work \cite{khalid2019optical}.

\section{Conclusion}

This work introduced MarineFormer, a transformer-based architecture leveraging spatial and temporal attention as a backbone for RL-based navigation of USVs in dynamic marine environments with spatially varying flows. A central contribution of this study is the proposal to incorporate flow field measurements into the state representation, which we demonstrate can have a significant impact on the solvability of the navigation problem in high-disturbance settings. We also show that the full benefit of this additional sensory input is only realized when effectively fused with ego and obstacle states. MarineFormer addresses this challenge through a novel attention-based architecture: spatial attention enables multimodal sensor fusion, while temporal attention, realized via a transformer encoder, models the evolving dynamics of the environment. Our results show that MarineFormer not only improves success rates and path efficiency over classical and learning-based baselines, but also generalizes well across RL frameworks and obstacle densities.

\bibliographystyle{IEEEtran}
\bibliography{Refs}

\end{document}